\documentclass[letterpaper]{article} 
\usepackage{aaai2026}  
\usepackage{times}  
\usepackage{helvet}  
\usepackage{courier}  
\usepackage[hyphens]{url}  
\usepackage{graphicx} 
\usepackage{natbib}  
\usepackage{caption} 
\usepackage{algorithm}
\usepackage{algorithmic}

\usepackage{newfloat}
\usepackage{listings}
\DeclareCaptionStyle{ruled}{labelfont=normalfont,labelsep=colon,strut=off} 
\floatstyle{ruled}
\newfloat{listing}{tb}{lst}{}
\floatname{listing}{Listing}
\usepackage{amsmath}
\usepackage{amsthm}  
\usepackage{graphicx}  
\usepackage{tcolorbox} 
\usepackage{multirow}
\usepackage{colortbl}
\usepackage{bbding}
\usepackage{array}  

\usepackage{amssymb,booktabs}
\usepackage{tabularx}
\usepackage{bbding}
\usepackage{pifont}
\usepackage{makecell}
\newcommand{\eg}{\textit{e.g.}}

\title{RemoteReasoner: Towards Unifying Geospatial Reasoning Workflow}

\author {
    Liang Yao\textsuperscript{\rm 1}\thanks{This work was done during a research internship at COWARobot. Liang Yao and Fan Liu contribute equally.},
    Fan Liu\textsuperscript{\rm 1}\thanks{Corresponding Author},
    Hongbo Lu\textsuperscript{\rm 2, 3},
    Chuanyi Zhang\textsuperscript{\rm 1}, \\
    Rui Min\textsuperscript{\rm 1}, 
    Shengxiang Xu\textsuperscript{\rm 4}, 
    Shimin Di\textsuperscript{\rm 4}, 
    Pai Peng\textsuperscript{\rm 3}\footnotemark[2]
}
\affiliations {
    \textsuperscript{\rm 1}Hohai University\\ 
    \textsuperscript{\rm 2}Shanghai Jiao Tong University\\
    \textsuperscript{\rm 3}COWARobot Co. Ltd.\\
    \textsuperscript{\rm 4}Southeast University\\
    fanliu@hhu.edu.cn, penmgpai\_sh@163.com
}

\usepackage{bibentry}

\begin{document}

\maketitle

\begin{abstract}
Remote sensing imagery presents vast, inherently unstructured spatial data, necessitating sophisticated reasoning to interpret complex user intents and contextual relationships beyond simple recognition tasks. In this paper, we aim to construct an Earth observation workflow to handle complex queries by reasoning about spatial context and user intent. As a reasoning workflow, it should autonomously explore and construct its own inference paths, rather than being confined to predefined ground‑truth sequences.
Ideally, its architecture ought to be unified yet generalized, possessing capabilities to perform diverse reasoning tasks through one model without requiring additional fine-tuning.
Existing remote sensing approaches rely on supervised fine-tuning paradigms and task‑specific heads, limiting both autonomous reasoning and unified generalization.
To this end, we propose RemoteReasoner, a unified workflow for geospatial reasoning. The design of RemoteReasoner integrates a multi-modal large language model (MLLM) for interpreting user instructions and localizing targets, together with task transformation strategies that enable multi-granularity tasks, including object-, region-, and pixel-level. 
In contrast to existing methods, our framework is trained with reinforcement learning (RL) to endow the MLLM sufficient reasoning autonomy. 
At the inference stage, our transformation strategies enable diverse task output formats without requiring task-specific decoders or further fine-tuning. Experiments demonstrated that RemoteReasoner achieves state-of-the-art performance across multi-granularity reasoning tasks. Furthermore, it retains the MLLM's inherent generalization capability, demonstrating robust performance on unseen tasks and categories. 

\end{abstract}

\begin{links}
    \link{Code}{https://github.com/1e12Leon/RemoteReasoner}
\end{links}

\section{Introduction}

Remote sensing analysis has evolved beyond simple object detection~\cite{liu2025boost,li2025rsvg} or classification~\cite{adegun2023review}, increasingly requiring stronger capabilities to support complex real-world decision-making.  
Although existing approaches~\cite{yao2025remotesam,liu2024remoteclip} are effective for multiple tasks, they struggle to interpret nuanced user intent and infer implicit spatial relationships for intricate queries. Reasoning~\cite{jaech2024openai,xu2025towards} bridges this gap by enabling systems to dynamically decipher the underlying goals of ambiguous or high-level instructions (\textit{e.g.}, \emph{``Find areas at risk of flooding near critical infrastructure after heavy rain"}). Such a geospatial workflow would conveniently accommodate varied geoscience applications~\cite{zhao2024artificial}.



\begin{figure}[t]
    \centering
    \includegraphics[width=0.95\linewidth]{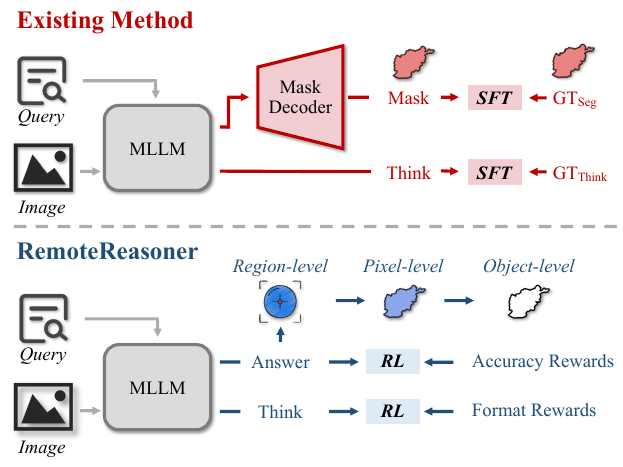}
    \caption{Comparison between Existing Frameworks and RemoteReasoner. Existing remote sensing reasoning approach (\eg, SegEarth-R1) requires SFT with annotated reasoning processes, and is limited to single-task outputs and task decoder. In contrast, our framework supports unsupervised reasoning and multi-granularity tasks.}
    \label{fig1}
\end{figure}

Numerous efforts~\cite{lai2024lisa,ren2024pixellm,chen2024sam4mllm} have explored reasoning tasks in the natural image domain. For instance, LISA~\cite{lai2024lisa} utilizes multi-modal large language models (MLLMs)~\cite{carolan2024review} for reasoning-driven segmentation tasks. However, these approaches rely on supervised fine-tuning (SFT) paradigms to train the models. Such frameworks constrain model autonomy, ultimately compromising generalization capabilities. This situation can be manifested in substantial performance degradation on out-of-distribution samples and catastrophic forgetting of general capabilities~\cite{liu2025seg}. Similarly, the remote sensing community has seen pioneering works~\cite{li2025segearth,zhang2023spatial,wang2024earthvqa}. SegEarth-R1~\cite{li2025segearth} introduces a geospatial pixel reasoning task that integrates a hierarchical MLLM and a mask decoder to generate both reasoning processes and segmentation outputs. However, it remains constrained by SFT paradigms and exclusively supports segmentation tasks.

These limitations motivate us to design a reasoning workflow that autonomously interprets user intent through self-directed thinking. It should also efficiently generate geospatial interpretations in diverse output formats, which are tailored for various downstream intelligence analysis.
Firstly, the requirement for self-thinking should be fulfilled through an open-ended learning paradigm. This aligns naturally with the emerging reinforcement learning (RL) frameworks~\cite{guo2025deepseek,shen2025vlm}. Rather than SFT approaches that demand strictly annotated reasoning processes, RL explores the model's self-evolution potential to cultivate reasoning capabilities. 
Secondly, to flexibly support multi-granularity tasks, we should avoid redundant task decoders and training through decision-level task transformation. 

Followed by these insights, we propose \textbf{RemoteReasoner}, a reasoning workflow that integrates a robust MLLM and a flexible inference strategy. As shown in Fig.~\ref{fig1}, building upon the \textit{pixel-level} reasoning task proposed in SegEarth-R1~\cite{li2025segearth}, RemoteReasoner further implements two novel tasks through autonomous reasoning: 1) \textit{region-level}: geospatial region reasoning, and 2) \textit{object-level}: geospatial contour reasoning. These three distinct tasks can be achieved through a single MLLM inference followed by a unified decision-level task transformation pipeline, significantly improving output efficiency for multi-granularity reasoning tasks.

During training, we utilize Group Relative Policy Optimization (GRPO)~\cite{shao2024deepseekmath} to fine-tune the reasoning model. Benefiting from our sophisticated reward function that balances output formats, localization accuracy, and quantity, RemoteReasoner demonstrates strong capabilities in autonomous reasoning and precise \textit{region-level} localization. This training approach maintains the inherent generalization capability of MLLM, thereby endowing it with the ability to recognize out-of-distribution categories. 
During inference, we design a task transformation pipeline inspired by RemoteSAM~\cite{yao2025remotesam}, which enables the model to perform \textit{object-} and \textit{pixel-level} reasoning tasks, eliminating redundant computations across granularities.

Extensive experimental results demonstrate that RemoteReasoner achieves state-of-the-art performance on multi-granularity reasoning tasks, evidencing its enhanced autonomous reasoning capabilities and superior generalization. For instance, RemoteReasoner achieves 3.67\% accuracy improvement over SegEarth-R1 in pixel-level geospatial reasoning tasks. On unseen tasks (\eg, Referring Segmentation), our framework significantly outperforms competing models (\eg, PixelLM, LISA) in performance.

The contributions are summarized as follows:

\begin{itemize}
    \item We propose RemoteResoner, a unified yet generalized reasoning workflow for Earth observation, which reveals that integrating pure RL and task transformation strategies could handle multi-granularity reasoning tasks.
    \item We introduce two novel geospatial reasoning tasks, region reasoning and contour reasoning. We also provide the corresponding precisely annotated datasets to facilitate the advancement of remote sensing research.
    \item Holistic evaluations demonstrate that RemoteReasoner achieves superior generalization capabilities and efficiency in handling multi-granularity reasoning tasks through a single forward pass.
\end{itemize}

\section{Related Work}
\label{sec:related_work}

\subsection{Reasoning in Large Language Models}
Recent efforts~\cite{jaech2024openai,guo2025deepseek,ren2024pixellm,lai2024lisa,li2025segearth} have shown that reasoning is central to aligning large language models (LLMs) with complex instructions. Models like OpenAI-o1~\cite{jaech2024openai} and DeepSeek-R1~\cite{guo2025deepseek} incorporate chain-of-thought(CoT)~\cite{wei2022chain} reasoning into their generation process, enabling flexible and interpretable multi-step inference. Both adopt reinforcement learning (RL)~\cite{kaelbling1996reinforcement,li2017deep,arulkumaran2017deep} to induce generalizable reasoning behaviors—OpenAI-o1 through RLHF and CoT monitoring, and DeepSeek-R1 via large-scale RL from scratch followed by multi-stage pretraining. These approaches exhibit strong reasoning capabilities but remain limited to natural language and vision tasks.

In remote sensing, models such as PixelLM~\cite{ren2024pixellm}, LISA~\cite{lai2024lisa}, and SegEarth-R1~\cite{li2025segearth} extend reasoning to spatial domains. While effective for pixel- or region-level inference, they often support narrow task types and fixed granularity. Notably, SegEarth-R1 relies on supervised fine-tuning, making it less robust to distribution shifts and less adaptable in open-ended settings. These limitations underscore the need for a more versatile geospatial framework. 

\begin{figure*}[t]
    \centering
    \includegraphics[width=1\linewidth]{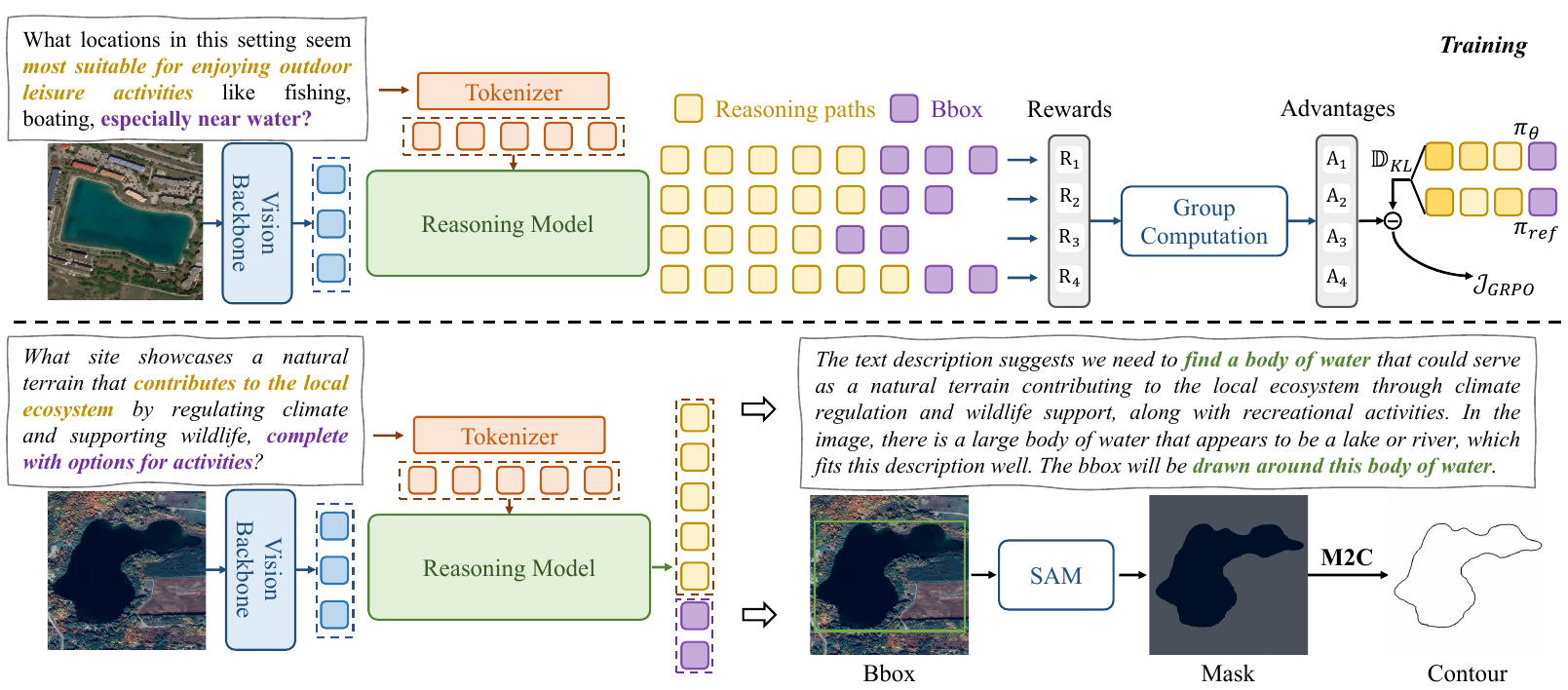}
    \caption{Overview of our RemoteReasoner. We utilize GRPO~\cite{shao2024deepseekmath} to explore the model's self-thinking capability. Then we design an inference workflow to perform multi-granularity reasoning tasks.}
    \label{overview}
\end{figure*}

\subsection{Remote Sensing Multi-modal Models}
Recent remote sensing multimodal models~\cite{liu2024remoteclip,zhang2024rs5m,hu2025rsgpt,ma2025unified,li2024show,li2024languag} demonstrate strong capabilities in handling diverse tasks across spatial granularities. Falcon~\cite{yao2025falcon} supports 14 vision-language tasks via multi-resolution encoding; GeoChat~\cite{kuckreja2024geochat} and LHRS-Bot~\cite{muhtar2024lhrs} incorporate spatial prompts and geographic priors for improved region-level grounding; RemoteSAM~\cite{yao2025remotesam} adopts a unified encoder for dense pixel-level segmentation; SkyEyeGPT~\cite{zhan2024skyeyegptunifyingremotesensing} reformulates task input/output as natural language for broad task generalization. While effective in geospatial understanding, these models lack robust reasoning capabilities, limiting performance under ambiguous or compositional prompts. While existing models demonstrate strong perception capabilities, they lack the reasoning ability to interpret complex prompts or adapt across granularities. 

\section{RemoteReasoner}
In this section, we introduce our method of building a workflow for multi-granularity geospatial reasoning tasks. The overall framework is represented in Fig.~\ref{overview}.

\subsection{Data}
The EarthReason~\cite{li2025segearth} dataset is proposed alongside SegEarth-R1, containing 6 questions per sample with an average length of 20.86 words. It comprises 2,371 training, 1,135 validation, and 1,928 testing samples. However, this dataset is limited to pixel tasks. To address this, we generate detection and contour annotations through Mask2Bbox (M2B)~\cite{liu2024remoteclip} and Mask2Contour (M2C)~\cite{miao2025prompting} strategies, respectively. Additionally, we design prompt templates combined with original questions and generated bounding boxes to form training samples. The segmentation and contour annotations are exclusively reserved for downstream testing evaluation.

\subsection{Reasoning Model} 
We utilize Qwen2.5-VL-7B-instruct~\cite{bai2025qwen2} as our reasoning model $\mathcal{F}_{\theta}$. Despite Qwen2.5-VL's superior performance in general-domain applications, direct adaptation to remote sensing imagery introduces domain knowledge discrepancy. Moreover, its reasoning capabilities require significant enhancement to address complex geospatial tasks. Therefore, we require fine-tuning it through reinforcement learning on remote sensing data. 

\subsection{Training} 
We utilize Group Relative Policy Optimization (GRPO)~\cite{shao2024deepseekmath} to train RemoteReasoner's reasoning model. Unlike traditional reinforcement learning methods like PPO~\cite{schulman2017proximal}, GRPO directly compares groups of candidate outputs without an additional critic model. Given a query $Q$, GRPO samples N candidate outputs {$o_1, o_2, ..., o_N$} from the policy $\pi_\theta$. Then it computes each output $o_i$ through a Reward function $R(Q, o_i)$.  To determine the relative quality of the outputs, GRPO further normalizes the obtained rewards and subsequently derives the advantage:

\begin{equation}
    A_i = \frac{r_i - mean\{r_1, r_2, ..., r_N\}}{std\{r_1, r_2, ..., r_N\}},
\end{equation}
where $A_i$ denotes the advantage of $o_i$ relative to other sampled outputs. GRPO aims to make the model generate outputs that obtain higher advantages, then utilize them to optimize the policy $\pi_\theta$:

\begin{equation}
    \begin{aligned}
    \mathcal{J}_{GRPO}(\theta) &= \mathbb{E}[\{o_i\}_{i=1}^N \sim \pi_{\theta_{old}}(q)] \\
    &\frac{1}{N}\sum_{i=1}^N\{min[c_1 \cdot A_i, c_2 \cdot A_i] \\
    & - \beta \mathbb{D}_{KL}[\pi_\theta || \pi_{ref}]\},
    \end{aligned}
\end{equation}
where, $c_1$ and $c_2$ can be formulated as follows:

\begin{equation}
\begin{aligned}
    c_1 &= \frac{\pi_\theta(o_i|q)}{\pi_{\theta_{\text{old}}}(o_i|q)}, \\
    c_2 &= \operatorname{clip}\left( \frac{\pi_\theta(o_i|q)}{\pi_{\theta_{\text{old}}}(o_i|q)}, 1 + \epsilon, 1 - \epsilon \right).
\end{aligned}
\end{equation}

To adopt GRPO for our reasoning model, we design a composite reward function to optimize localization precision, object count accuracy, and output format. The total reward $R$ combines 3 critical components:

\begin{itemize}
    \item \textbf{Accuracy reward} measures bounding box alignment using Intersection-over-Union (IoU) with optimal Hungarian matching. For predicted boxes $B = \{b_k\}_{k=1}^K$ and ground truth $G = \{g_j\}_{j=1}^J$:

    \begin{equation}
    R_{\text{IoU}} = \frac{1}{|G|} \sum_{j=1}^{|G|} \underset{k}{\text{max}} \left[ \frac{|b_k \cap g_j|}{|b_k \cup g_j|} \right].
    \end{equation}

    \item  \textbf{Count reward} penalizes deviations in object quantity prediction using an exponential decay function:

    \begin{equation}
    R_{\text{count}} = 
    \begin{cases} 
    1 & (J = 0) \land (K = 0) \\
    0 & (J = 0) \land (K > 0) \\
    \exp\left(-2 \cdot \frac{|K - J|}{J}\right) & J > 0
    \end{cases},
    \end{equation}
    where $J = |G|$ and $K = |B|$ denote ground-truth and predicted counts respectively. The exponential term ($-2\cdot$) creates a smooth penalty gradient that becomes stricter as relative count error increases.   

    \item \textbf{Format reward}  $R_{\text{format}}$ checks whether RemoteReasoner's outputs follow the specified format, returning 1 or 0 based on compliance. It is the JSON-style with reasoning paths in the $<$think$>$...$</$think$>$ tag and bboxes in the $<$answer$>$[[$x_1$,$y_1$,$x_2$,$y_2$]...]$</$answer$>$ tag. 
\end{itemize}

Finally, the overall reward function is:

\begin{equation}
    R = R_{\text{IoU}} + R_{\text{count}} + R_{\text{format}}.
\end{equation}

\subsection{Inference Workflow} 
After training, the reasoning model itself generates bounding boxes (bbox) indicating target regions. To further enhance the workflow and enable both pixel-level and object-level reasoning tasks, we need to leverage established tools. Specifically, we first input the bbox and its centroid coordinates to the SAM2~\cite{ravi2024sam} to generate pixel-level masks. Subsequently, morphological operations are applied to analyze the mask images, extracting the outermost polygonal vertices to obtain target contours. The mathematical formulation of this process is presented as follows:

Given a remote sensing image $I \in \mathbb{R}^{H \times W \times 3}$ and a textual query $Q$, we define a multi-stage workflow for multi-task reasoning.


\begin{itemize}
    \item \textbf{Geospatial Region Reasoning:} 
    A vision-language model $\mathcal{F}_{\theta}$ processes the image-query pair $(I, Q)$ to jointly produce:
    \begin{equation}
        (T, b) = \mathcal{F}_{\theta}(I, Q),
    \end{equation}
    where $T$ denotes the textual reasoning trace, and $b = [x_{\text{min}}, y_{\text{min}}, x_{\text{max}}, y_{\text{max}}]$ is the normalized bounding box of the target region.
    
    \item \textbf{Geospatial Pixel Reasoning:}
    The bounding box $b$ and its center points $v$ serve as prompts for the Segment Anything Model (SAM) $\mathcal{S}$. SAM generates a binary mask $M \in \{0,1\}^{H \times W}$ isolating the target:
    \begin{equation}
        M = \mathcal{S}\big(I\,;\,b,v\big),
    \end{equation}
    where $M_{(i,j)} = 1$ indicates pixel $(i,j)$ belongs to the target.
    
    \item \textbf{Geospatial Contour Reasoning:}
    The target contour $C$ is derived  $M$ via morphological boundary detection:
    \begin{equation}
        C = \Gamma\big(M\big) = \{(x_k, y_k)\}_{k=1}^K,
    \end{equation}
    where $\Gamma(\cdot)$ extracts the polygonal vertices $\{(x_k, y_k)\}$ of the mask's external boundary.
\end{itemize}

\section{Experiments}

\begin{table}[t]
\centering

\begin{tabular}{c|cc|cc}
\toprule
\multirow{2}{*}{Method} & \multicolumn{2}{c|}{$cIoU$} & \multicolumn{2}{c}{$gIoU$} \\
\cline{2-5}
                        & Val         & Test       & Val         & Test       \\
\hline  \rowcolor{gray!20}                      
\multicolumn{5}{l}{\textit{SFT Methods}}                                               \\
\hline
LISA                  & 57.39       & 59.10      & 61.04       & 60.88      \\
PixelLM               & 57.79       & 59.22      & 57.94       & 60.01      \\
PSALM              & 62.03       & 64.61      & 66.61       & 68.30      \\
SegEarth-R1             & 64.13       & 68.25      & 68.60       & 70.75      \\
\hline \rowcolor{gray!20}
\multicolumn{5}{l}{\textit{RL Method}}                                                 \\
\hline \rowcolor{blue!15}
\textbf{RemoteReasoner}       & \textbf{67.80}      &  \textbf{69.13}         & \textbf{69.02}       &  \textbf{70.96}         \\
\bottomrule
\end{tabular}
\caption{Geospatial Pixel Reasoning Results on EarthReason. It is worth noting that RemoteReasoner was not trained directly utilizing its segmentation labels.}
\label{pixel}
\end{table}

\begin{table}[t]
\centering

\begin{tabular}{c|cc|cc}
\toprule
\multirow{2}{*}{Method} & \multicolumn{2}{c|}{$Acc@0.5$} & \multicolumn{2}{c}{$gIoU$} \\
\cline{2-5}
                        & Val         & Test       & Val         & Test       \\
\hline
DeepSeek-VL2-tiny                & 12.08       & 12.67      & 17.51       & 18.62     \\
GeoChat                  & 10.10       & 8.89      & 12.57       & 11.44      \\
Qwen2.5-VL-7B                   &    41.21    &  45.82    &   38.77     &   41.80  \\
\hline \rowcolor{blue!15}
\textbf{RemoteReasoner}       & \textbf{66.51}     &   \textbf{68.11}         & \textbf{67.04}       &    \textbf{69.29}       \\
\bottomrule
\end{tabular}
\caption{Geospatial Region Reasoning Results. }
\label{region}
\end{table}

\begin{figure}[t]
    \centering
    \includegraphics[width=1\linewidth]{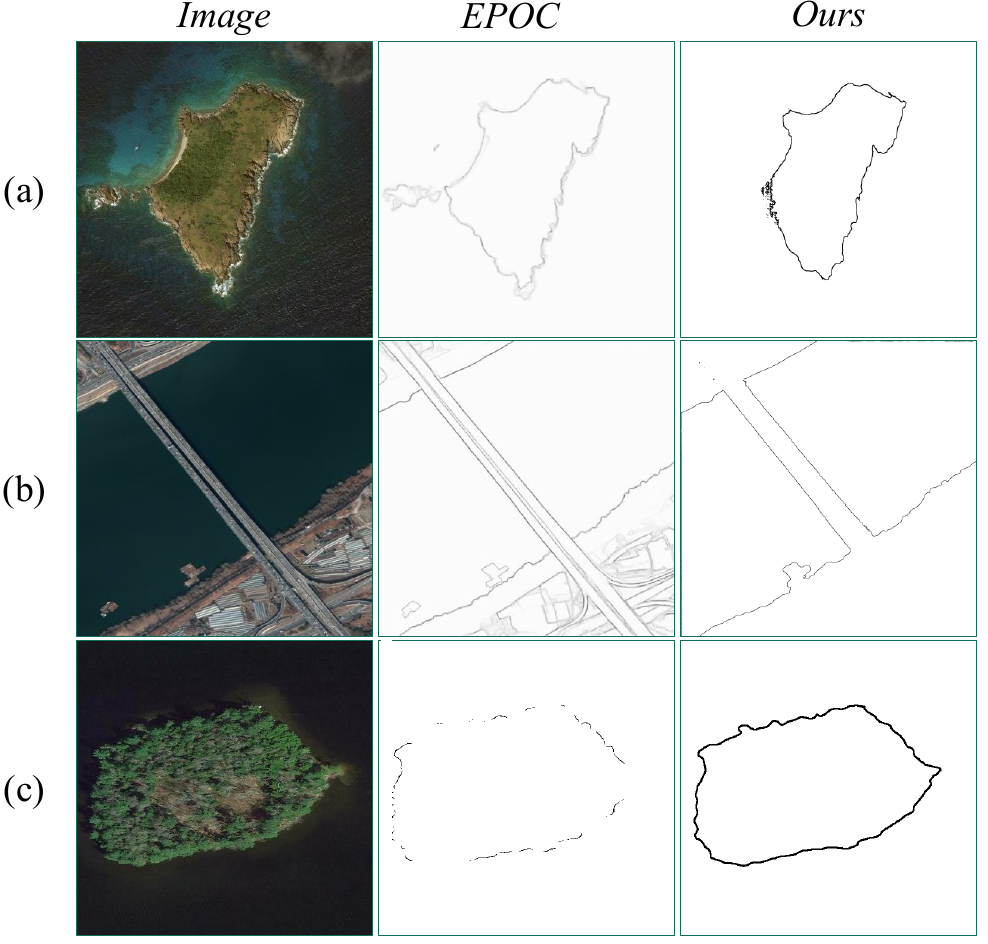}
    \caption{Geospatial Contour Reasoning Results. We select EPOC~\cite{chen2024subobject} for comparison.}
    \label{ablation:contour}
\end{figure}

\begin{table*}[t]
\centering
\begin{tabular}{c|cc|cc|cc|cc|cc|cc}
\toprule
\multirow{2}{*}{Method} & \multicolumn{2}{c|}{$F1@1$$\uparrow$} & \multicolumn{2}{c|}{$F1@3$$\uparrow$} & \multicolumn{2}{c|}{$Recall@1$$\uparrow$} & \multicolumn{2}{c|}{$Recall@3$$\uparrow$} & \multicolumn{2}{c|}{$ASD$$\downarrow$} & \multicolumn{2}{c}{$HD$$\downarrow$} \\
\cline{2-13}
                        & Val         & Test       & Val         & Test     & Val         & Test       & Val         & Test   & Val         & Test       & Val         & Test      \\
\hline
EPOC                  & 0.10       & 0.10      & 0.19       & 0.20  &0.14 &0.15  &0.28 &0.29 & 60.07      & 64.38      & 243.98      & 255.81     \\
\hline \rowcolor{blue!15}
\textbf{RemoteReasoner}       & \textbf{0.46}      &  \textbf{0.44}         & \textbf{0.55}       &  \textbf{0.58}   &\textbf{0.40} &\textbf{0.47}  &\textbf{0.53} & \textbf{0.64} & \textbf{41.50}      &  \textbf{45.20}         & \textbf{160.56}       &  \textbf{165.11}        \\
\bottomrule
\end{tabular}
\caption{Geospatial Contour Reasoning Results on EarthReason.}
\label{contour}
\end{table*}

\subsection{Dataset and Metrics}
\textbf{Dataset.} We evaluate on the validation and test sets of EarthReason~\cite{li2025segearth}. Pixel-level tasks directly utilize the original annotations, while region-level and object-level tasks are transformed using M2B~\cite{liu2024remoteclip} and M2C~\cite{miao2025prompting} annotation formats based on the original labels, respectively. Additionally, we employ the RSVG~\cite{zhan2023rsvg} test set for Visual Grounding evaluation and the RRSISD~\cite{liu2024rotated} test set for Referring Expression Segmentation tasks, with their training sets explicitly excluded from model training. For unseen category evaluation, we manually selected 100 images covering 10 distinct unseen categories from the RemoteSAM-270K~\cite{yao2025remotesam} dataset.

\textbf{Metrics.} Following previous works~\cite{yao2025remotesam,li2025segearth}, we adopt $gIoU$, $cIoU$, and $Acc@0.5$ as region- and pixel-level evaluation metrics. For contour reasoning, we followed~\cite{jiao2025roipoly} to employ F1 score with 1-pixel and 3-pixel dilated kernel ($F1@1$ \& $F1@3$) as evaluation metrics, and incorporate two geometric error measures: Average Symmetric Distance ($ASD$) and Hausdorff Distance ($HD$).

\subsection{Implementations Details}

We employ GRPO~\cite{shao2024deepseekmath} to fine-tune the Qwen2.5-VL-7B-Instruct~\cite{bai2025qwen2} model using LoRA~\cite{hu2022lora} (rank=8, alpha=16, targeting all linear layers). Training is conducted for 24 epochs on 8 × NVIDIA L20 GPUs with a global batch size of 512 (achieved via per-device batch size 8 and gradient accumulation steps 8), using the AdamW~\cite{loshchilov2017decoupled} optimizer with a learning rate of 1e-6, bf16 precision, and DeepSpeed~\cite{rasley2020deepspeed} Zero3 for memory efficiency.

\subsection{Main Results}

\subsubsection{Geospatial Pixel Reasoning}

We evaluate RemoteReasoner's performance on pixel-level reasoning tasks using the EarthReason dataset. We compare against state-of-the-art general-domain methods (LISA~\cite{lai2024lisa}, PixelLM~\cite{ren2024pixellm}, PSALM~\cite{zhang2024psalm}) and the remote sensing-domain SegEarth-R1~\cite{li2025segearth}, with results summarized in Tab.~\ref{pixel}. Experimental results demonstrate that despite not directly utilizing pixel-level labels during training, our approach remarkably surpasses SFT-based methods that leverage both pixel annotations and reasoning trajectories. For instance, RemoteReasoner achieves 10.41\% higher $cIoU$ than LISA on the validation set. Compared to the best-performing SegEarth-R1, we observe a 3.67\% improvement. The primary reason lies in the superior robustness of reinforcement learning (RL) frameworks compared to supervised fine-tuning (SFT) paradigms. Even without direct exposure to pixel-level annotations, our model effectively utilizes bounding boxes to prompt the Segment Anything Model (SAM), achieving precise segmentation results through this indirect yet efficient interaction mechanism.

\subsubsection{Geospatial Region Reasoning}

This section validates RemoteReasoner's performance on region-level reasoning tasks, with results presented in Tab.~\ref{region}. Compared to the remote sensing multi-modal foundation model GeoChat~\cite{kuckreja2024geochat}, RemoteReasoner demonstrates at least a 54.47\% improvement in $Acc@0.5$. This significant gain stems from GeoChat's weak reasoning capabilities, which fail to interpret ambiguous queries. Similarly, RemoteReasoner achieves over 22.29\% improvement against general-purpose MLLMs, \textit{e.g.,} Qwen-2.5VL-7B~\cite{bai2025qwen2} and DeepSeek-VL2-tiny (3B)~\cite{guo2025deepseek}. Although these MLLMs possess basic reasoning abilities, their insufficient domain-specific knowledge in remote sensing leads to suboptimal performance on geospatial region-level reasoning tasks.

\subsubsection{Geospatial Contour Reasoning}
This section validates the contour extraction performance of RemoteReasoner. We select EPOC~\cite{chen2024subobject} as a comparative baseline. Since EPOC lacks reasoning capabilities, we crop the bounding boxes (bbox) generated by RemoteReasoner and input them into EPOC to obtain contours, as shown in Fig.~\ref{ablation:contour}. As EPOC is class-agnostic, it extracts contours for all objects within the region (\textit{e.g.}, Fig.~\ref{ablation:contour} (a) and (b)). In contrast, RemoteReasoner's contours are derived from segmentation masks, resulting in tighter alignment with target boundaries. Furthermore, under low-resolution conditions (Fig.~\ref{ablation:contour} (c)), EPOC produces blurry or disconnected contours, whereas our method maintains sharp and closed boundaries.

Quantitative results (as shown in Tab.~\ref{contour}) demonstrate that RemoteReasoner surpasses EPOC by at least 0.34 in both F1 scores, with significantly lower distance errors compared to EPOC. This performance gap can be attributed to EPOC's tendency to extract extraneous contours and its failure to handle indistinct target boundaries effectively.

\begin{table}[t]
\centering
\renewcommand\arraystretch{1.02}
\setlength{\tabcolsep}{3.2mm}{
\begin{tabular}{c|c|cc}
\toprule
Task                                   & Method            & $Acc@0.5$ & $gIoU$  \\
\hline
\multirow{9}{*}{VG}                    &\multicolumn{3}{l}{\cellcolor{gray!20}\textit{Trained}}        \\ 
                                       \cline{2-4}
                                       & MGVLF             & 76.78   & 68.04     \\
                                       & Falcon            & 56.87   & -     \\
                                       & SkyEyeGPT         & 70.50   & -     \\ 
                                       & LHRS-Bot          & 73.45   & -     \\
                                       \cline{2-4}
                                       &\multicolumn{3}{l}{\cellcolor{gray!20}\textit{Zero-shot}}        \\ 
                                       \cline{2-4}
                                       & MiniGPT-V2        & 46.64   & -     \\
                                       & Qwen-VL-Chat      & 44.76   & -     \\ 
                                       & \textbf{RemoteReasoner} & \textbf{50.76}   & \textbf{49.85} \\
\bottomrule                                       
Task                                & Method            & $cIoU$    & $gIoU$  \\

\hline                      
\multirow{8}{*}{RES}                &\multicolumn{3}{l}{\cellcolor{gray!20}\textit{Trained}}                   \\
\cline{2-4} 
                                       & GeoGround         & -       & 60.50 \\
                                       & SegEarth-R1       & 67.56   & 66.40 \\
                                       \cline{2-4}
                                       &\multicolumn{3}{l}{\cellcolor{gray!20}\textit{Zero-shot}}        \\ 
                                       \cline{2-4}
                                       & LISA              & -       & 26.78 \\
                                       & PixelLM           & -       & 31.65 \\
                                       & NEXT-Chat         & -       & 24.98 \\
                                       & \textbf{RemoteReasoner} & \textbf{54.29}   & \textbf{50.97 } \\
\bottomrule                                       
\end{tabular}
}
\caption{Performance on unseen tasks. VG and RES are Visual Grounding and Referring Expression Segmentation. }
\label{VG_RES}
\end{table}

\begin{figure}[t]
    \centering
    \includegraphics[width=0.98\linewidth]{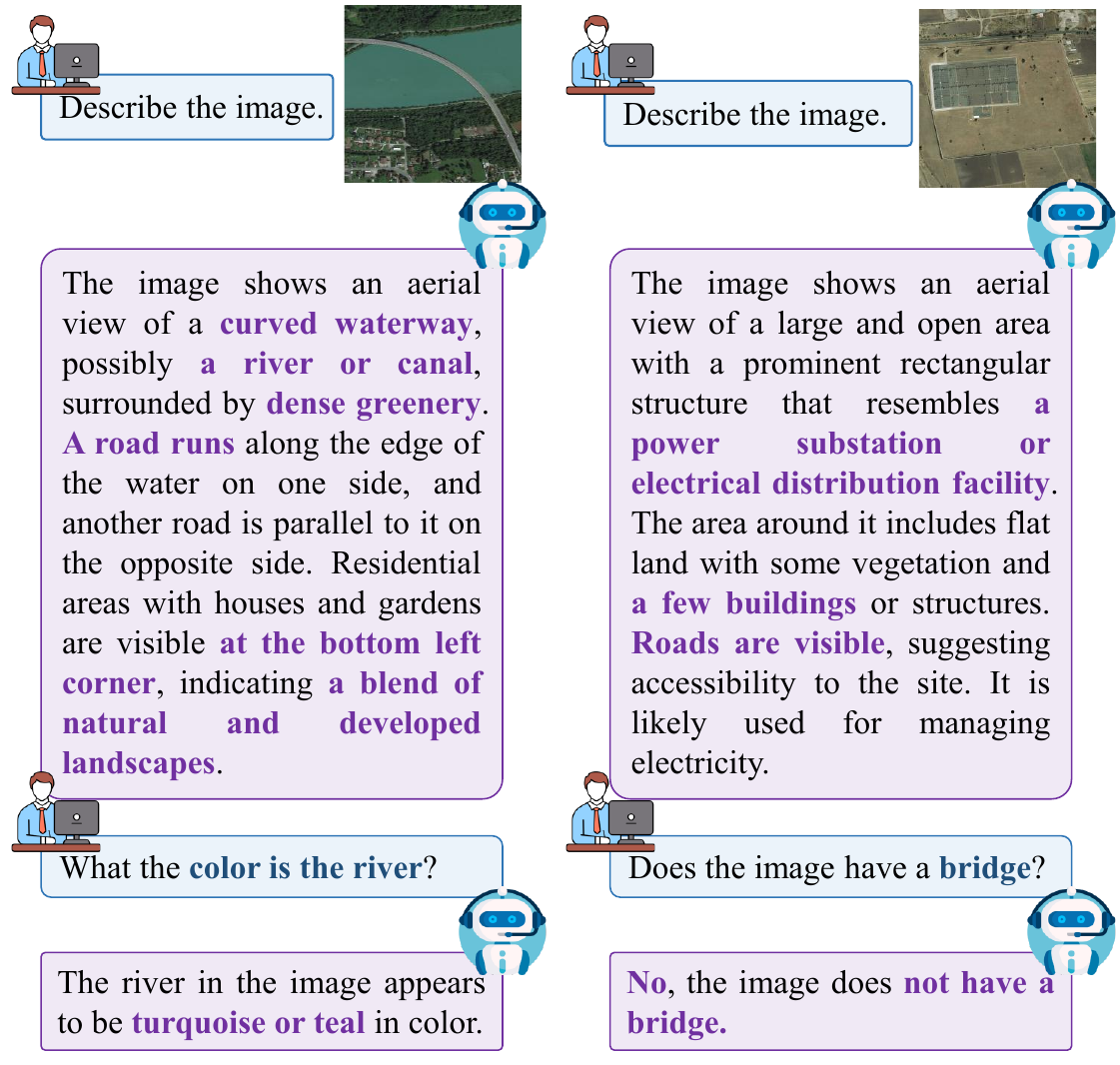}
    \caption{Quantitative results of other tasks (Image Captioning \& VQA). }
    \label{qa}
\end{figure}

\begin{table}[t]
\centering
\setlength{\tabcolsep}{1mm}{
\begin{tabular}{c|ccccc}
\toprule
\textbf{Category} & Car    & Ship   & Field   & Apron & Grass      \\
\hline
$gIoU$     & 21.56  & 50.03  & 36.72   & 54.09 & 38.27      \\
\toprule
\textbf{Category} & Trucks & Hangar & Parking & Roof  & Warehouses \\
\hline
$gIoU$     & 25.58  & 58.34  & 30.52   & 52.85 & 41.00   \\
\bottomrule
\end{tabular}
}
\caption{Out-of-distribution category recognition performance. We manually select 10 novel classes that are absent in EarthReason.}
\label{OOD}
\end{table}

\begin{table}[t]
\setlength{\tabcolsep}{2.8mm}{
\begin{tabular}{c|c|cc}
\toprule
Output Format              & Training Method & $cIoU$  & $gIoU$  \\
\hline
\multirow{2}{*}{Text2Mask}      & SFT             & 22.01 & 22.97 \\
                           & RL              & 18.74 & 19.21 \\
                           
\hline                           
\multirow{2}{*}{Bbox2Mask} & SFT             & 60.25 & 62.19 \\

                           & RL              & \textbf{69.13} & \textbf{70.96}\\
\bottomrule
\end{tabular}
}
\caption{Ablation on different pixel-level output formats. Text2Mask denotes LLM-generated textual polygon.}
\label{text}
\end{table}

\begin{figure*}[t]
    \centering
    \includegraphics[width=1\linewidth]{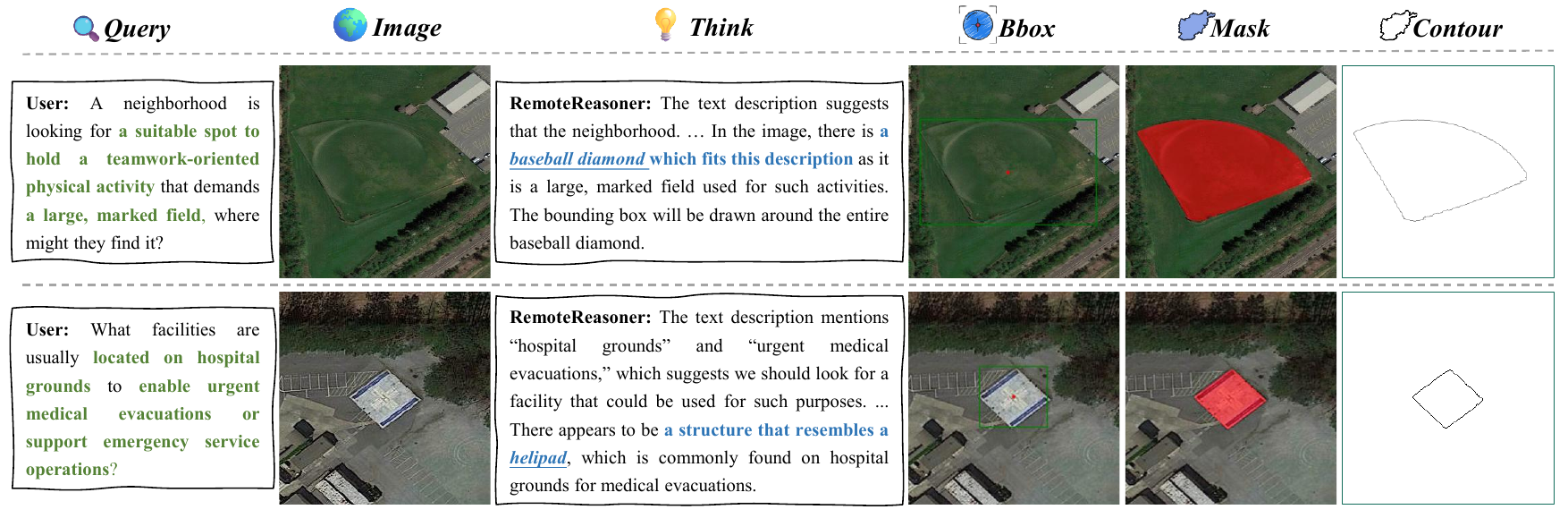}
    \caption{Qualitative Results. Given an image and its corresponding implicit query, RemoteReasoner autonomously identifies the user-intended target category through reasoning and accurately executes visual-centric tasks across three granularities.}
    \label{further}
\end{figure*}

\begin{figure}[t]
    \centering
    \includegraphics[width=1\linewidth]{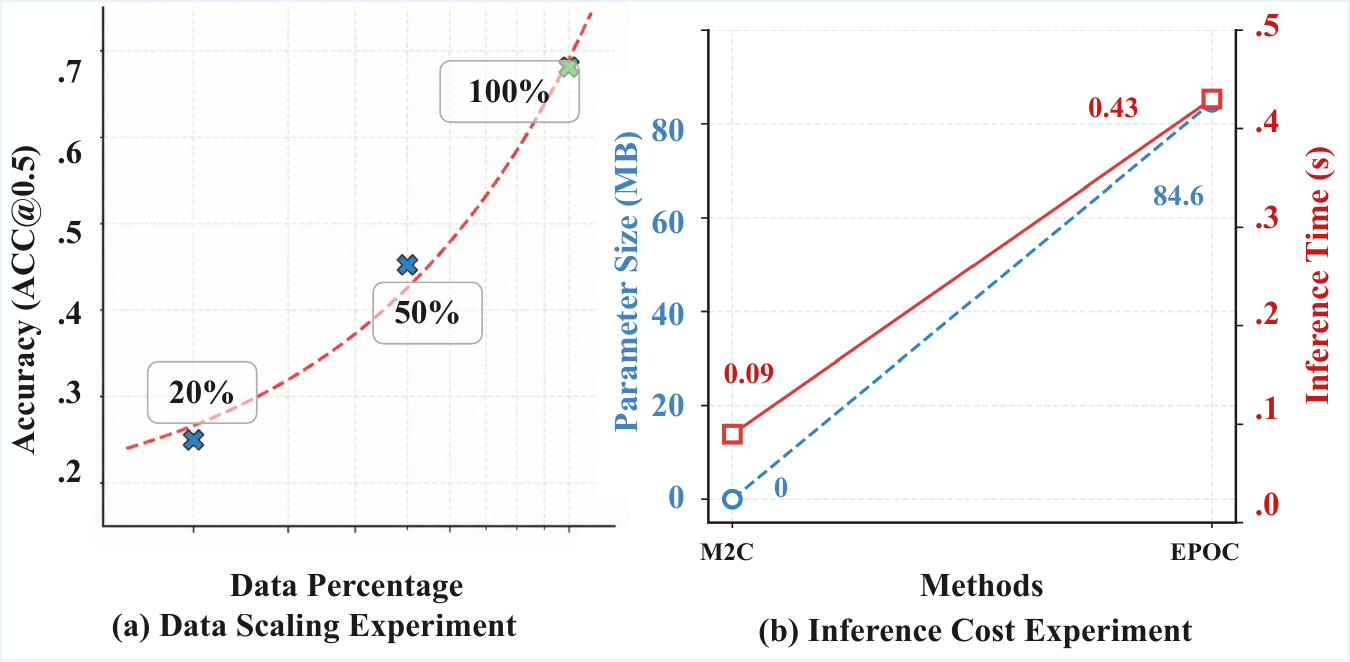}
    \caption{(a) Data scaling experiment. We observed that the model performance increases as the data scales exponentially. (b) Inference cost experiment. Mask2Coutour method achieved significantly faster inference speed. }
    \label{time}
\end{figure}


\subsection{Further Analysis}
After reinforcement learning training, can the reasoning model maintain the original capabilities of the MLLM? To explore this, we evaluate RemoteReasoner's extrinsic capabilities, including its ability to handle unseen task types and recognize out-of-distribution categories.

\subsubsection{Can RemoteReasoner Do Other Tasks? }
We evaluate RemoteReasoner on Visual Grounding and Referring Expression Segmentation tasks, with results summarized in Tab.~\ref{VG_RES}. Comparative experiments are conducted against state-of-the-art LLM-based approaches, including general-domain methods (MiniGPT-V2~\cite{zhu2023minigpt}, LISA~\cite{lai2024lisa}, PixelLM~\cite{ren2024pixellm}) and remote sensing-specific frameworks (Falcon~\cite{yao2025falcon}, LHRS-Bot~\cite{muhtar2024lhrs}, GeoGround~\cite{zhou2024geoground}). Notably, all remote sensing methods were trained on task-specific data, while RemoteReasoner's zero-shot performance achieves 80\% of their supervised results. Under identical zero-shot settings, our framework achieves gains of 4.12\% on Visual Grounding and 19.32\% on Referring Expression Segmentation tasks. 

Furthermore, as the workflow's core component is a multi-modal large language model, we verify preservation of its inherent capabilities through multi-turn dialogue evaluation (Fig.~\ref{qa}). The model first performs image captioning, then accurately answers content-based questions about the image. Additional results are provided in the supplementary material. These results demonstrate our framework's strong generalization capabilities with RL.

\subsubsection{Can RemoteReasoner Recognize Other Category? }
Given that the EarthReason~\cite{li2025segearth} dataset contains only 28 categories, which is significantly insufficient to cover common remote sensing targets. Therefore, we further investigate RemoteReasoner's recognition capability for unseen categories. We manually select 100 images spanning 10 novel categories from RemoteSAM-270K~\cite{yao2025remotesam} that are absent in EarthReason for evaluation. As represented in Tab.~\ref{OOD}, RemoteReasoner demonstrates robust recognition of out-of-distribution categories with consistent accuracy. It demonstrates that our RL strategy avoids overfitting to category distributions in the training data. However, the model still exhibits insufficient capabilities in detecting small objects (\eg, cars), indicating a direction for future research.

\subsubsection{Qualitative Analysis}
The qualitative results of the three supported tasks in our workflow are presented in Fig.~\ref{further}. In the first sample row, the user seeks a venue for group sports, while a baseball field nearly blends into the background in the image. RemoteReasoner correctly reasons and locates the target, simultaneously yielding its segmentation mask and contour. The second row shows a relatively blurred image of a medical helipad. Based on its function of providing emergency rescue services, RemoteReasoner accurately locates and segments the target area along with its contour. These visualizations indicate that our method can adapt to challenging reasoning tasks.


\subsection{Ablation Studies}


\subsubsection{Different Pixel-level Output Formats}
Recently, RemoteSAM~\cite{yao2025remotesam} validates that segmentation, as a fundamental visual output unit, enables bottom-up transformation to other vision tasks. Therefore, if LLMs could directly generate masks in textual representations, the architectural design holds promise for greater unification. To explore this hypothesis, we conducted experiments using such formulations, as summarized in Tab.~\ref{text}. However, textual outputs prove ill-suited for this predictive task, with both SFT and RL strategies exhibiting significant performance degradation compared to visual-based approaches. In contrast, directly outputting bounding boxes and then generating masks through SAM significantly outperforms textual mask representations (22.01 vs. 60.25). When incorporating RL strategies, the accuracy further improves from 60.25 to 69.13. Based on these findings, we adopt the Mask2Contour approach in our workflow design.

\subsubsection{Data Scaling Experiment}
To validate data impact on model performance, we evaluate it by training with 20\%, 50\%, and 100\% subsets of the full training data. As illustrated in Fig.~\ref{time} (a), the results demonstrate that model performance improves with increasing data scale. The full dataset (100\%) achieves the best performance, indicating the current data scale contains no redundancy. These findings suggest that geospatial reasoning tasks retain untapped potential that can be unlocked through larger data scales.

\subsubsection{Inference Cost Ablation}

We further evaluate inference speed across different contour extraction strategies, as shown in Fig.~\ref{time} (b). The Mask2Contour approach directly applies morphological operations without introducing additional parameters, achieving significantly faster processing speed (0.09s per image). In contrast, EPOC, as a deep learning model with 84.6MB parameters, demonstrates inferior inference speed (0.43s per image). This comparison highlights the efficiency advantages of our proposed workflow.

\section{Conclusion}
\label{sec:conclusion}

We present RemoteReasoner, a novel workflow advancing autonomous geospatial reasoning. Overcoming limitations of SFT-based approaches, it leverages RL strategy to cultivate robust reasoning capabilities while preserving MLLM generalization. Its core innovation is a unified inference pipeline enabling efficient multi-granularity task transformation (pixel-, region-, and object-level) from a single MLLM output, eliminating redundant per-task decoders. RemoteReasoner achieves SOTA results, including $>$ 15\% and 30\% accuracy gains in region- and object-level reasoning, and demonstrates superior generalization across diverse, complex queries. It provides a flexible foundation for downstream geospatial intelligence requiring nuanced intent interpretation and multi-format outputs.

\section*{Acknowledgments}

This work was supported in part by the National Natural Science Foundation of China under Grant 62372155 and Grant 62302149, in part by Basic Research Program of Jiangsu under Grant BK20250188, in part by the Research Funds of Jiangsu Hydraulic Research Institute under Grant 2025z065, and in part by Changzhou Science and Technology Bureau Project
No. 20231313, and in part by Research and Application of the Next-Generation General-Purpose Intelligent Robot Brain (Robo-GPT).

\bibliography{aaai2026}

\end{document}